\documentclass[11pt,a4paper]{article}

\usepackage[preprint]{acl}
\usepackage{times}
\usepackage{latexsym}
\usepackage[T1]{fontenc}
\usepackage[utf8]{inputenc}
\usepackage{microtype}
\usepackage{graphicx}
\usepackage{amsmath}
\usepackage{booktabs}
\usepackage{array}
\usepackage{url}

\title{Prompt Framing Distorts Count-Based Evaluation of LLM Error Detection: Evidence from Numeric Anchoring}
\author{Dekun Yang \\
  Zhejiang University \\
  \texttt{pauliyangwork@gmail.com}}
\date{}

\begin{document}
\maketitle

\begin{abstract}
Count-based F1 can improve even when an error detector does not localize errors more accurately. ErrorBench tests this evaluation failure with six LLMs, five prompt conditions, and 3,690 retained responses to 123 CoNLL-2014 passages. Changing a supplied count from $N-2$ to $N+2$ moves the reported count by 3.28--3.74 errors for the four GPT and Claude systems, or 82--94\% of the four-error prompt difference; the two Gemini systems move by 0.14--0.21. In an 83-passage corrected-text replication, changing the prompt from Blind to one that states the true count raises Count-F1 by 0.22 on average but raises an ERRANT-based, two-reference edit $F_{0.5}$ by only 0.04. Paired bootstrap intervals place the difference between these changes above zero for five of six models. A description-derived M2-style analysis on all 123 passages shows the same mismatch. Because the prompt supplies the target used by the count metric, high count agreement is not evidence of better localization. Evaluations of LLM proofreading should avoid supplying expected error counts and pair count metrics with edit- or span-aware scores.
\end{abstract}

\section{Introduction}
Large language models (LLMs) increasingly serve as proofreaders, code reviewers, and factual checkers. Evaluations of these systems sometimes reduce error detection to a single question: does the reported error count match the reference count? A system can then receive an F1 near $1.0$ even when it poorly identifies \emph{which} errors occur. We call this discrepancy the count--span gap. A single sentence in the prompt can widen the gap enough to change a count-only evaluation. The problem lies in exposing expected counts during evaluation, not in the relative quality of any one model family.

Comparable cues already appear in error-correction evaluations. MEDIQA-CORR 2024 defines each clinical note as containing one or no medical error, and its GPT-4 baseline states that constraint in the prompt \citep{ben-abacha-etal-2024-overview}. A system submitted to the same task injects an upstream model's predicted error span into the correction prompt as a hint \citep{gema-etal-2024-edinburgh-clinical}. Both designs serve their stated task, but they put count or location information into a downstream model's input. If an evaluation rewards agreement with that information, its score can mix detection quality with prompt compliance.

We isolate the count component by changing only the number stated in the prompt while holding the passage and task fixed. We use anchoring as a behavioral description: humans often adjust insufficiently from salient reference values \citep{tversky1974judgment,epley2006anchoring}, and related effects have been reported for LLMs \citep{macmillanscott2024irrationality}. Our design does not separate anchoring from output imitation or instruction following. For evaluation, the relevant test is whether a supplied count improves count agreement without a commensurate improvement in localization.

We ask whether a supplied count shifts the number of errors reported, whether a true-count prompt improves Count-F1 more than edit quality, and how the response varies across model families. The paper makes three contributions. \textbf{(1)} A direct comparison of conflicting high and low count cues shows large reported-count shifts for the sampled GPT and Claude systems. \textbf{(2)} An 83-passage corrected-text replication uses ERRANT-based edit extraction and two CoNLL-2014 references. Blind$\to$Anchored Count-F1 rises by 0.22 on average, compared with 0.04 for edit $F_{0.5}$; the paired-bootstrap contrast is positive at the 95\% level for five of six models. \textbf{(3)} A 123-passage analysis provides Count Bias (CB), a descriptive Anchoring Sensitivity Index (ASI), and description-derived M2-style diagnostics. ErrorBench audits an evaluation protocol; it is not a model ranking or a test of internal mechanism.

\section{Related Work}
\subsection{LLM Evaluation and Error Detection}
A growing body of work evaluates LLMs on structured proofreading and fact-checking tasks. \citet{fang2023chatgptgec} report competitive grammatical-error-correction results for ChatGPT in some settings, together with systematic over-correction. BEA-2019 \citep{bryant2019bea}, CoNLL-2014 \citep{ng2014conll}, and the broader GEC literature \citep{bryant2023survey} also show why count agreement alone is an incomplete measure of correction quality. In scientific reviewing, pilot studies find that targeted error-finding prompts can work better than holistic review prompts \citep{liu2023reviewergpt}; related work studies AI-assisted peer review and review corpora \citep{checco2021aipeerreview,dycke2023nlpeer}.

This sensitivity to how the error-finding request is framed raises a further question: what happens when the prompt supplies a specific expected count?

\subsection{Prompt Sensitivity and Sycophancy}
LLM outputs vary with prompt wording, instruction order, and demonstration selection \citep{zhao2021calibrate,lu2022fantastically,min2022rethinking}. Sycophancy is especially relevant here: models may agree with a user's stated belief even when it is false \citep{perez2023discovering,sharma2024sycophancy}. \citet{perez2023discovering} observe this behavior across several knowledge domains, indicating that a prompt can override information available to the model. Positional and social biases also affect LLM-as-judge assessments \citep{ye2025llmasjudge}. We study a narrower case in which the prompt supplies a number and the model estimates an error count.

\subsection{Numeric Cues and Evaluation Contamination}
The anchoring effect is well established in cognitive psychology \citep{tversky1974judgment}. Under the anchoring-and-adjustment account, people begin with a salient reference value and stop adjusting before reaching an independent estimate \citep{epley2006anchoring}. LLM studies have reported related effects in reasoning, forecasting, and numerical-judgment tasks \citep{macmillanscott2024irrationality,nguyen2024anchoring,huang2025anchoring,lou2026anchoring}; recent evidence further relates susceptibility to model confidence and post-training \citep{owusu2026anchoring}. Our setting differs in one important respect: the supplied number is also the target used by the count metric. We therefore focus on prompt-induced count movement and score distortion rather than the internal mechanism.

\section{Methodology}
\subsection{Dataset Construction}
We build ErrorBench from the CoNLL-2014 Shared Task data \citep{ng2014conll}. Consecutive sentences are grouped into non-overlapping four-sentence windows, and passage counts and gold edits come from Annotator 0. The initial stratified sample contained 143 windows with 3--7 errors. A post-collection integrity audit aligned the ordered M2 sentences to the 50 \texttt{<DOC>} blocks in the official SGML source and found that 20 sampled windows crossed an essay boundary. We excluded those windows without inspecting model outputs, leaving 123 passages: 25, 28, 23, 26, and 21 in the 3--7 error buckets. The original sample and exclusion list remain in the release for auditability.

Each passage stores raw text, its Annotator-0 count, M2 categories, and token-indexed annotations. The initial corrected-text subset contained 100 passages; the same audit excluded 17 boundary-crossing windows, leaving 83. ERRANT 3.0.0 extracts edits from model corrections \citep{bryant2017errant}. We then apply exact tuple matching against Annotator 0 and a two-reference sensitivity rule using Annotators 0 and 1 (Section~\ref{sec:edit_replication}).

\subsection{Prompt Conditions}
We design five prompt conditions as our primary independent variable, shown in Table~\ref{tab:conditions}. All conditions share a fixed system prompt instructing the model to respond using a structured format (``ERROR N: [description]'' followed by ``TOTAL ERRORS FOUND: N'') so that outputs can be parsed automatically. Temperature is fixed at 0 to reduce sampling variation; this does not make remote model inference deterministic.

\begin{table*}[t]
\centering
\small
\caption{Five prompt conditions used in ErrorBench, varying the prior information provided about error count.}
\label{tab:conditions}
\begin{tabular}{>{\raggedright\arraybackslash}p{0.14\textwidth} >{\raggedright\arraybackslash}p{0.29\textwidth} >{\raggedright\arraybackslash}p{0.19\textwidth} >{\raggedright\arraybackslash}p{0.24\textwidth}}
\toprule
Condition & Prompt framing & Prior info given & Cue tested \\
\midrule
Blind & Does this text have errors? & None & No-count baseline \\
Informed & This text has errors---find them. & Errors exist & Existence cue \\
Anchored & This text has exactly $N$ errors---find them. & True count $N$ & True-count cue \\
Mislead-Over & This text has exactly $N+2$ errors---find them. & Over-count ($N+2$) & Inflated count cue \\
Mislead-Under & This text has exactly $\max(1,N-2)$ errors---find them. & Under-count ($\max(1,N-2)$) & Deflated count cue \\
\bottomrule
\end{tabular}
\end{table*}

For the Misleading conditions, we set $M=N+2$ (Mislead-Over) and $M=\max(1,N-2)$ (Mislead-Under), where $N$ is the true error count. The supplied value therefore differs from the truth by exactly 2 in both directions.

\subsection{Models}
We evaluate GPT-4o and GPT-5.4 \citep{openai2024gpt4o,openai2026gpt54}, Claude Haiku 4.5 and Claude Sonnet 4.6 \citep{anthropic2025haiku45,anthropic2026sonnet46}, and Gemini 2.5 Flash and Gemini 3.1 Pro Preview \citep{deepmind2025gemini25flash,google2026gemini31}. The labels follow the requested model identifiers at an OpenAI-compatible proxy; cross-family comparisons are descriptive. The initial collection planned $143\times6\times5=4{,}290$ cells. All post-audit main analyses use the complete $123\times6\times5=3{,}690$ grid and a keep-last-success rule for retries. The main prompt uses an 800-token output limit; the corrected-text replication uses 1,400 tokens.

\subsection{Evaluation Metrics}
ErrorBench is a controlled stress test. For passage $i$, let $\hat N_i$ be the reported count and $N_i$ the reference count. Count Bias is $\mathrm{CB}_i=\hat N_i-N_i$. For condition $c$, $\mathrm{ASI}_i(c)=|\mathrm{CB}_i(c)-\mathrm{CB}_i(\mathrm{Blind})|/N_i$; we report its mean and SD over parseable condition--Blind pairs. We call this descriptive measure the Anchoring Sensitivity Index (ASI); the name refers to the prompt manipulation and does not identify an internal mechanism. Count-F1 uses count overlap without span matching: $\mathrm{TP}_i=\min(\hat N_i,N_i)$, $\mathrm{FP}_i=\max(0,\hat N_i-N_i)$, and $\mathrm{FN}_i=\max(0,N_i-\hat N_i)$. We average passage-level Count-F1. An unparseable count receives Count-F1 0; CB and paired count analyses use parseable pairs and report their effective sample size $n$. We also compare Mislead-Over directly with Mislead-Under. Because all retained passages have $N_i\geq3$, these prompts differ by exactly four errors.

The main span diagnostic is description-derived and M2-style; it is not the official M2 scorer. We parse quoted source/correction fragments from each error description, localize them in the passage, and form $(\mathrm{sentence},\mathrm{start},\mathrm{end},\mathrm{correction})$ tuples. Unlocalized descriptions are false positives in the primary analysis. Repository-local strict, detection, and overlap matching produce corpus micro $F_{0.5}$ scores; Appendix~\ref{app:m2_full} gives definitions and a sensitivity analysis. The corrected-text replication avoids description parsing: ERRANT extracts predicted and reference edits, followed by our documented exact-matching and two-reference selection rules. Missing corrected text is scored as an empty prediction. Because Count-F1 and edit $F_{0.5}$ are not commensurate scales, inference focuses on their changes under the same prompt contrast, not only on their raw difference.

For reproducibility, the appendix reports the prompts, model identifiers, decoding settings, and scoring variants needed to audit the experiment. The full code, prompts, and raw model outputs will be released upon publication in an anonymized research repository; all figures are generated from the passage-level model-output records using the same analysis scripts described in the appendix.

\section{Experiments and Results}
\subsection{Main Results}
Table~\ref{tab:cb} reports Count Bias and Count-F1. GPT-5.4 follows the Mislead-Over count with zero variance (CB $=+2$), while Claude H.4.5 is closest to the true count in the Anchored condition (CB $=-0.016$, SD $=0.180$). Near-perfect Anchored Count-F1 is partly built into the prompt. Blind behavior differs sharply: GPT-5.4 over-reports by 7.3 errors on average, Claude S.4.6 by 2.9, and Claude H.4.5 by 0.2. The two Gemini systems instead undercount across conditions. This is lower count responsiveness in this protocol, not evidence of greater general robustness.

\begin{table*}[t]
\centering
\footnotesize
\caption{Count Bias (mean and SD over parseable counts) and mean passage-level Count-F1. Count-F1 uses all 123 passages, assigning 0 to an unparseable count. CB uses $n=123$ except for Gemini 3.1 ($n=117,118,120,120,121$ in table order).}
\label{tab:cb}
\begin{tabular}{llccc}
\toprule
Model & Condition & CB Mean & CB SD & Count-F1 \\
\midrule
GPT-4o & Blind & +1.244 & 2.338 & 0.814 \\
GPT-4o & Informed & +2.146 & 5.572 & 0.811 \\
GPT-4o & Anchored & +0.033 & 0.254 & 0.996 \\
GPT-4o & Mislead-Over & +2.000 & 0.000 & 0.821 \\
GPT-4o & Mislead-Under & -1.740 & 0.818 & 0.737 \\
\addlinespace[1pt]
GPT-5.4 & Blind & +7.268 & 3.105 & 0.582 \\
GPT-5.4 & Informed & +7.886 & 3.252 & 0.563 \\
GPT-5.4 & Anchored & +0.146 & 0.786 & 0.988 \\
GPT-5.4 & Mislead-Over & +2.000 & 0.000 & 0.821 \\
GPT-5.4 & Mislead-Under & -1.285 & 2.201 & 0.721 \\
\addlinespace[1pt]
Claude H.4.5 & Blind & +0.163 & 1.826 & 0.858 \\
Claude H.4.5 & Informed & +0.715 & 1.831 & 0.853 \\
Claude H.4.5 & Anchored & -0.016 & 0.180 & 0.999 \\
Claude H.4.5 & Mislead-Over & +1.976 & 0.271 & 0.821 \\
Claude H.4.5 & Mislead-Under & -1.634 & 0.934 & 0.754 \\
\addlinespace[1pt]
Claude S.4.6 & Blind & +2.862 & 3.712 & 0.758 \\
Claude S.4.6 & Informed & +2.667 & 3.666 & 0.770 \\
Claude S.4.6 & Anchored & -0.016 & 0.384 & 0.993 \\
Claude S.4.6 & Mislead-Over & +1.854 & 0.796 & 0.819 \\
Claude S.4.6 & Mislead-Under & -1.683 & 0.862 & 0.741 \\
\addlinespace[1pt]
Gemini 2.5 & Blind & -2.146 & 1.721 & 0.702 \\
Gemini 2.5 & Informed & -2.252 & 1.607 & 0.707 \\
Gemini 2.5 & Anchored & -2.374 & 1.596 & 0.696 \\
Gemini 2.5 & Mislead-Over & -2.398 & 1.551 & 0.694 \\
Gemini 2.5 & Mislead-Under & -2.610 & 1.303 & 0.647 \\
\addlinespace[1pt]
Gemini 3.1 & Blind & -2.752 & 1.666 & 0.582 \\
Gemini 3.1 & Informed & -2.737 & 2.040 & 0.582 \\
Gemini 3.1 & Anchored & -2.833 & 1.463 & 0.599 \\
Gemini 3.1 & Mislead-Over & -2.892 & 1.340 & 0.589 \\
Gemini 3.1 & Mislead-Under & -3.033 & 1.147 & 0.546 \\
\bottomrule
\end{tabular}
\end{table*}

\begin{figure*}[t]
\centering
\includegraphics[width=0.96\textwidth]{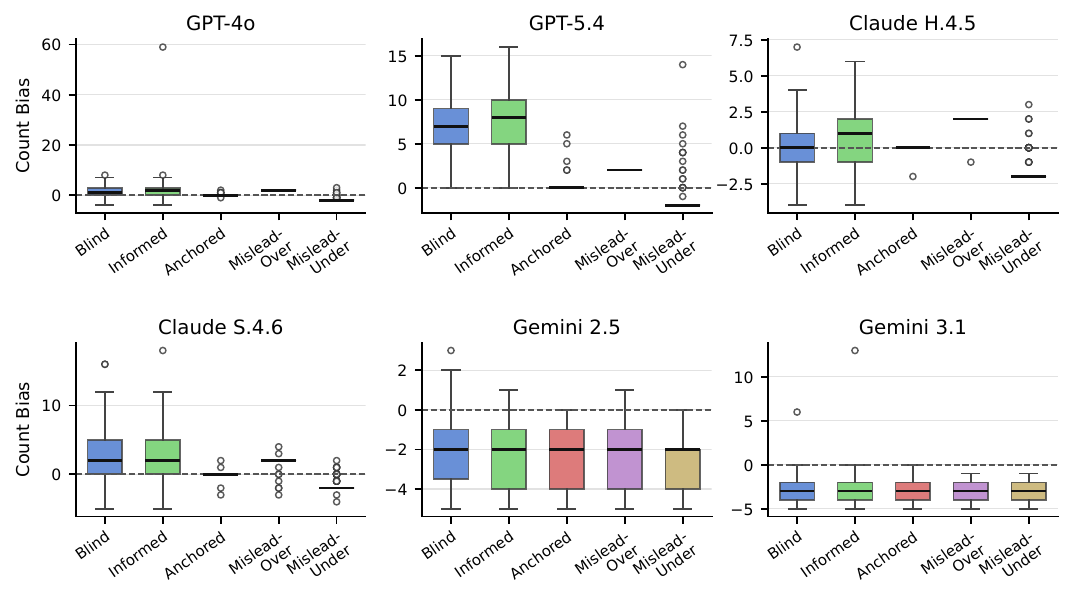}
\caption{Count Bias distributions across prompt conditions. Boxes show the median and IQR; whiskers extend to $1.5\times$IQR. When the prompt supplies a count, GPT-5.4 and both Claude models track it closely, whereas both Gemini models continue to undercount.}
\label{fig:cb_boxplot}
\end{figure*}

\begin{figure*}[t]
\centering
\includegraphics[width=0.96\textwidth]{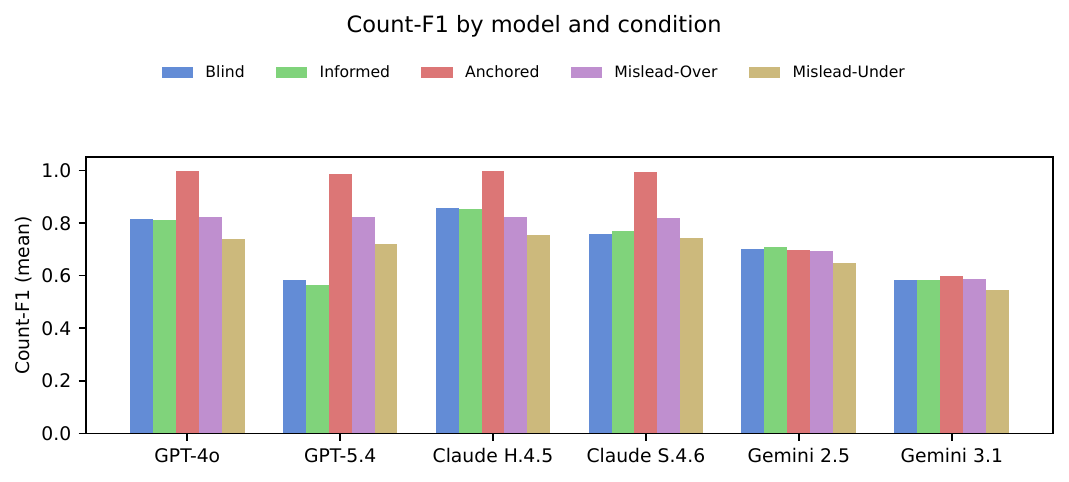}
\caption{Count-F1 by condition and model. The Anchored condition achieves near-perfect Count-F1 for GPT-5.4, Claude H.4.5, and Claude S.4.6 because the prompt supplies the target count; both Gemini systems show little count response under a persistent undercount prior.}
\label{fig:count_f1_bar}
\end{figure*}

\subsection{Response to Conflicting Count Cues}
\begin{figure*}[t]
\centering
\includegraphics[width=0.78\textwidth]{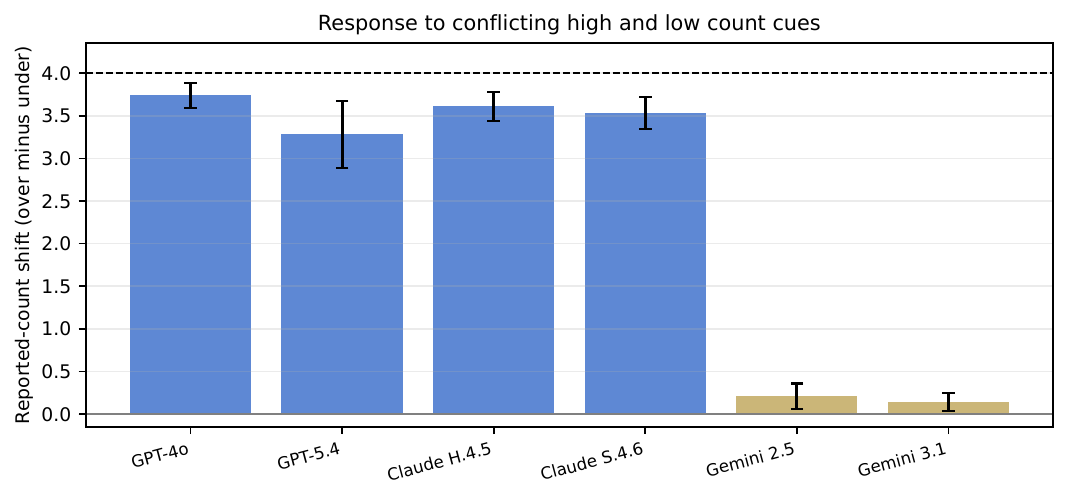}
\caption{Paired reported-count shift from Mislead-Under to Mislead-Over. Error bars are 95\% t intervals. The dashed line is the four-error difference between the prompts. Gemini 3.1 uses 118 parseable pairs; all other models use 123.}
\label{fig:anchor_response}
\end{figure*}

Mislead-Over and Mislead-Under differ only in the supplied count, by four errors. Their paired contrast therefore avoids using the true-count condition as the sole evidence of a count-cue response. The mean reported-count shift is 3.740 for GPT-4o, 3.285 for GPT-5.4, 3.610 for Claude H.4.5, and 3.537 for Claude S.4.6 (Figure~\ref{fig:anchor_response}). These values are 82--94\% of the prompt difference. The shifts for Gemini 2.5 and Gemini 3.1 are 0.211 and 0.144. All six paired shifts differ from zero after BH correction over these six tests ($q<.01$), although the Gemini effect sizes are small. The high count is repeated exactly in 93--100\% of GPT/Claude outputs and 0\% of Gemini outputs.

\subsection{Anchoring Sensitivity Index}
Table~\ref{tab:asi} relates each prompted condition to Blind. GPT-5.4 has the largest value: Mislead-Under ASI $=1.928$. Claude S.4.6 reaches 1.073, compared with 0.751 for GPT-4o and 0.456 for Claude H.4.5. Gemini 2.5 remains at or below 0.187 and Gemini 3.1 at or below 0.120. Because low ASI co-occurs with undercounting, it is not a clean estimate of resistance to count cues.

\begin{table*}[t]
\centering
\small
\caption{Anchoring Sensitivity Index (ASI, passage-level mean $\pm$ SD) by model and prompt condition.}
\label{tab:asi}
\begin{tabular}{lccc}
\toprule
Model & ASI (Anchored) & ASI (Mislead-Over) & ASI (Mislead-Under) \\
\midrule
GPT-4o & 0.490 $\pm$ 0.517 & 0.417 $\pm$ 0.360 & 0.751 $\pm$ 0.609 \\
GPT-5.4 & 1.625 $\pm$ 0.949 & 1.234 $\pm$ 0.873 & 1.928 $\pm$ 1.069 \\
Claude H.4.5 & 0.304 $\pm$ 0.292 & 0.427 $\pm$ 0.259 & 0.456 $\pm$ 0.406 \\
Claude S.4.6 & 0.772 $\pm$ 0.845 & 0.598 $\pm$ 0.662 & 1.073 $\pm$ 0.980 \\
Gemini 2.5 & 0.187 $\pm$ 0.218 & 0.170 $\pm$ 0.216 & 0.170 $\pm$ 0.221 \\
Gemini 3.1 & 0.083 $\pm$ 0.249 & 0.075 $\pm$ 0.239 & 0.120 $\pm$ 0.299 \\
\bottomrule
\end{tabular}
\end{table*}

\begin{figure*}[t]
\centering
\includegraphics[width=0.72\textwidth]{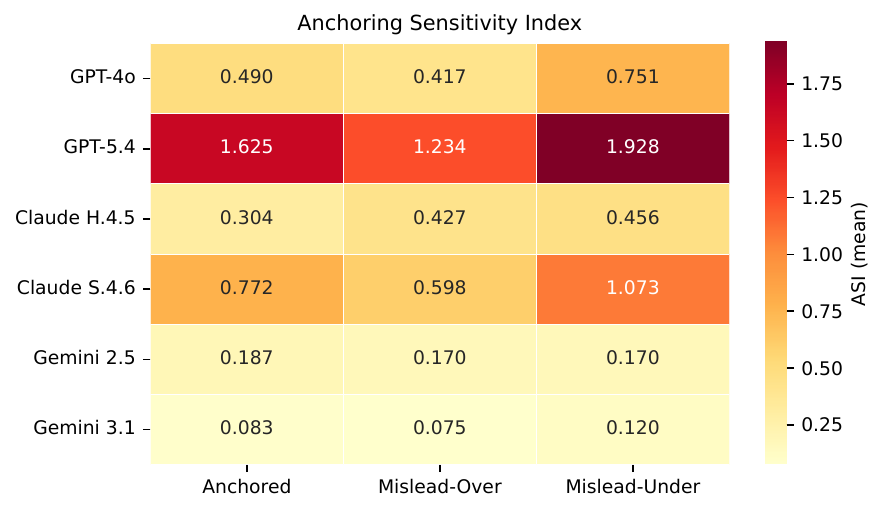}
\caption{Mean ASI over the 123-passage analysis set (115 parseable pairs for each displayed Gemini 3.1 condition except Mislead-Over, $n=114$).}
\label{fig:asi_heatmap}
\end{figure*}

\subsection{Statistical Significance}
We compare each condition with Blind using paired passage-level t-tests and one Benjamini--Hochberg correction over the 24 tests. Mislead-Under is significant for every model ($q\leq.0033$). Mislead-Over is significant for the four GPT/Claude systems ($q\leq.0043$), but not for Gemini 2.5 or 3.1 ($q=.057$ and $.131$). The true-count Anchored contrast is significant for GPT-4o, GPT-5.4, and Claude S.4.6, but not Claude H.4.5 or either Gemini model. Informed differs from Blind only for GPT-5.4 ($\Delta$CB $=+0.618$, $q=.012$) and Claude H.4.5 ($+0.553$, $q<10^{-5}$).

Paired effect sizes give the scale of these shifts. For GPT-5.4, $d_z$ is $-2.54$ for Mislead-Under, $-2.32$ for Anchored, and $-1.70$ for Mislead-Over. Mislead-Under is also large for GPT-4o ($-1.34$), Claude H.4.5 ($-1.05$), and Claude S.4.6 ($-1.21$). The Gemini values are smaller ($|d_z|\leq0.35$ for Gemini 2.5 and $\leq0.30$ for Gemini 3.1). Appendix~\ref{app:effects} reports all paired effects and effective sample sizes.

\subsection{Count Agreement and Description-derived Edits}
Table~\ref{tab:spanf1} compares Count-F1 with the description-derived M2-style overlap score. The raw gap is not a scale-free effect because the metrics have different matching rules. The within-metric change is more informative: averaged over models, Blind$\to$Anchored raises Count-F1 by 0.163 and overlap $F_{0.5}$ by 0.059. For GPT-5.4, Count-F1 changes from 0.582 to 0.988 while overlap $F_{0.5}$ stays near 0.20. For Claude H.4.5 the changes are $0.858\to0.999$ and $0.434\to0.461$. Under strict matching, the largest Anchored count--edit gap is 0.954 (Appendix~\ref{app:m2_full}).

\begin{table*}[t]
\centering
\footnotesize
\caption{Mean passage Count-F1, description-derived M2-style overlap $F_{0.5}$, and their diagnostic difference. The raw difference compares metrics with different matching rules.}
\label{tab:spanf1}
\begin{tabular}{llccc}
\toprule
Model & Condition & Count-F1 & M2 $F_{0.5}$ & Difference \\
\midrule
GPT-4o & Blind & 0.814 & 0.364 & 0.450 \\
GPT-4o & Anchored & 0.996 & 0.375 & 0.621 \\
GPT-4o & Mislead-Over & 0.821 & 0.326 & 0.495 \\
GPT-4o & Mislead-Under & 0.737 & 0.436 & 0.301 \\
\addlinespace[1pt]
GPT-5.4 & Blind & 0.582 & 0.201 & 0.381 \\
GPT-5.4 & Anchored & 0.988 & 0.202 & 0.786 \\
GPT-5.4 & Mislead-Over & 0.821 & 0.170 & 0.651 \\
GPT-5.4 & Mislead-Under & 0.721 & 0.276 & 0.445 \\
\addlinespace[1pt]
Claude H.4.5 & Blind & 0.858 & 0.434 & 0.424 \\
Claude H.4.5 & Anchored & 0.999 & 0.461 & 0.538 \\
Claude H.4.5 & Mislead-Over & 0.821 & 0.408 & 0.413 \\
Claude H.4.5 & Mislead-Under & 0.754 & 0.498 & 0.256 \\
\addlinespace[1pt]
Claude S.4.6 & Blind & 0.758 & 0.277 & 0.481 \\
Claude S.4.6 & Anchored & 0.993 & 0.509 & 0.484 \\
Claude S.4.6 & Mislead-Over & 0.819 & 0.462 & 0.357 \\
Claude S.4.6 & Mislead-Under & 0.741 & 0.559 & 0.182 \\
\addlinespace[1pt]
Gemini 2.5 & Blind & 0.702 & 0.269 & 0.433 \\
Gemini 2.5 & Anchored & 0.696 & 0.253 & 0.443 \\
Gemini 2.5 & Mislead-Over & 0.694 & 0.255 & 0.439 \\
Gemini 2.5 & Mislead-Under & 0.647 & 0.287 & 0.360 \\
\addlinespace[1pt]
Gemini 3.1 & Blind & 0.582 & 0.246 & 0.336 \\
Gemini 3.1 & Anchored & 0.599 & 0.342 & 0.257 \\
Gemini 3.1 & Mislead-Over & 0.589 & 0.299 & 0.290 \\
Gemini 3.1 & Mislead-Under & 0.546 & 0.398 & 0.148 \\
\bottomrule
\end{tabular}
\end{table*}

\subsection{Corrected-text Edit Replication}
\label{sec:edit_replication}
The 83-passage replication asks for a full corrected passage, uses ERRANT to extract edits, and scores exact edit tuples against both CoNLL annotators. Our two-reference sensitivity rule selects the higher-$F_{0.5}$ reference separately for each sentence; this is not the official M2 max-match scorer. Every model--condition cell contains all 83 passages. One missing call and any missing corrected-text blocks are empty predictions.

\begin{figure*}[t]
\centering
\includegraphics[width=0.86\textwidth]{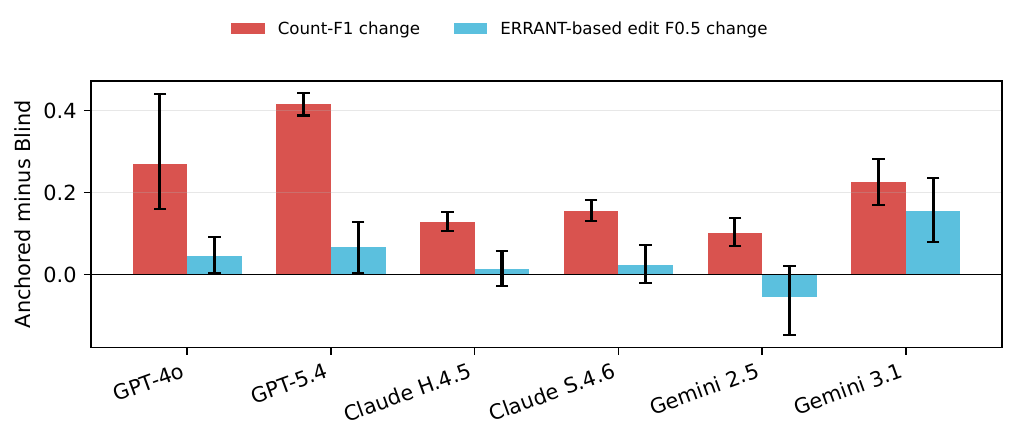}
\caption{Paired bootstrap on 83 passages. Bars show the mean Blind$\to$Anchored change with 95\% percentile intervals. The contrast between Count-F1 and edit $F_{0.5}$ is above zero for five models; the Gemini 3.1 contrast crosses zero.}
\label{fig:bootstrap_ci}
\end{figure*}

Across models, Blind$\to$Anchored raises corpus Count-F1 by 0.218 on average and two-reference edit $F_{0.5}$ by 0.042. In a paired passage bootstrap (1,000 draws, seed 42), the difference between these metric changes is positive at the 95\% level for GPT-4o, GPT-5.4, both Claude systems, and Gemini 2.5 (Figure~\ref{fig:bootstrap_ci}). Gemini 3.1 has a positive point estimate but its interval crosses zero. The bootstrap treats the missing call and unparseable counts as failures. Appendix~\ref{app:errant} gives the model-level bootstrap contrasts and full Path-B condition scores.

\subsection{Case Studies}
\label{sec:cases}
Case A gives the clearest qualitative failure. On \texttt{conll\_0238} ($N=4$), GPT-4o reports only the valid \emph{theirselves}$\to$\emph{themselves} correction under Blind. Under Mislead-Over ($M=6$), it repeats that correction and adds five unsupported claims: four nonexistent spaces before punctuation and an unnecessary article before \emph{depression}. It reaches the supplied count by inventing errors rather than recovering the other three gold edits. Appendix~\ref{app:cases} gives the passage and verbatim output, together with two contrasting cases. Under a low count, Claude H.4.5 drops from 11 reported errors to two on \texttt{conll\_0160} ($N=4$); Gemini 2.5, by contrast, still reports two errors when given a high count of five on \texttt{conll\_0201} ($N=3$).

\section{Discussion}
The experiment identifies evaluation contamination, not a cognitive mechanism. For four systems, the stated number largely determines the reported count even when two conflicting counts are applied to identical passages. GPT-5.4, for example, over-reports under Blind (CB $=+7.27$) but repeats every Mislead-Over count. The pattern resembles anchoring-and-adjustment \citep{epley2006anchoring}, but output imitation and instruction following are also plausible. The evaluation result is the same under any of these accounts: the prompt supplies part of the answer scored by Count-F1 while edit quality improves far less.

The within-family differences are sizable. Mislead-Under ASI is 1.928 for GPT-5.4 and 0.751 for GPT-4o; the corresponding values are 1.073 for Claude S.4.6 and 0.456 for Claude H.4.5. Gemini 2.5 and Gemini 3.1 remain below 0.187 and 0.120. Their low ASI coincides with under-reporting and frequent corrected-text omissions, so the data do not separate low count-cue responsiveness from a prior toward fewer reported edits.

The MEDIQA-CORR examples show that count and span hints already occur in at least one clinical error-correction setting. We do not claim that such cues are widespread. The risk arises whenever a review prompt contains a target count, including document review or code auditing. The 83-passage corrected-text study shows that the present result is not limited to description parsing or one reference; transfer to other domains remains an empirical question.

Absolute scores should not be compared with published GEC systems because our passages, prompts, missing-output policy, and two-reference rule differ. The within-model prompt contrast is the relevant evidence: Anchored Count-F1 is at least 0.906 for every model, whereas two-reference edit $F_{0.5}$ remains within 0.068 of Blind in five of six cases.

\subsection{Practical Recommendations for LLM Error-Detection Evaluation}
The results suggest four safeguards for LLM proofreading and document review. \textbf{Use blind prompts by default:} do not disclose an expected error count unless the evaluation is specifically testing count-cue sensitivity. \textbf{Treat count metrics as auxiliary:} Count-F1 and exact-count accuracy measure coarse calibration, not localization. \textbf{Report span-aware metrics:} M2, ERRANT, or an equivalent localization score should accompany count results. \textbf{Monitor exact count matching:} deployed systems should track how often model outputs reproduce counts supplied by users, templates, or upstream tools.

\section{Conclusion}
Supplying an error count can distort count-based evaluation. Conflicting high and low counts move four systems by 82--94\% of the count difference. In the corrected-text replication, Blind$\to$Anchored changes Count-F1 by 0.218 on average and two-reference edit $F_{0.5}$ by 0.042; the paired contrast is supported for five of six models. Count agreement obtained by stating the reference count should therefore not be read as better error localization. ErrorBench measures a weakness in the evaluation protocol, not overall model quality or an internal cognitive mechanism.

\section*{Limitations}
\paragraph{Span scoring.} The 123-passage analysis extracts edits from short descriptions and applies custom M2-style matching, not the official M2 scorer. The stronger corrected-text analysis covers 83 passages and uses ERRANT for edit extraction followed by custom exact matching. Extending that protocol to all 123 passages would reduce uncertainty.

\paragraph{Missing outputs and instruction compliance.} Gemini 2.5 and Gemini 3.1 omit a corrected-text block in 22--68 of 83 trials per condition. These are empty edit predictions, so their absolute edit scores combine localization and format compliance. The main log also contains 19 unparseable Gemini 3.1 counts; these receive Count-F1 0 and are excluded from CB. One Path-B call is missing and is scored as a failure.

\paragraph{References and matching.} Reference counts and supplied prompt values use Annotator 0. The corrected-text sensitivity analysis also uses Annotator 1, selecting the higher-$F_{0.5}$ reference per sentence. This permissive local rule is not the official M2 max-match procedure and can raise absolute scores.

\paragraph{Single inference and fixed decoding.} Each cell has one remote inference at temperature 0. The output limit is 800 tokens in the main run and 1,400 in Path B. Temperature 0 reduces but does not eliminate run-to-run variation, and a shared setting may depart from provider-specific recommendations.

\paragraph{Limited count range.} The Mislead conditions use only $N\pm2$. They establish a response to this perturbation but not its dose--response curve.

\paragraph{Post-collection boundary audit and domain scope.} We discovered after collection that 20 of 143 main windows and 17 of 100 Path-B windows crossed source-document boundaries. The exclusion rule used the official SGML boundaries and was applied without reference to outputs, but it reduced the sample and left unequal count buckets. CoNLL-2014 learner English may not represent other review domains.

\paragraph{Model-specific confounds.} Gemini's low ASI coincides with persistent undercounting. The experiment cannot separate low responsiveness to count cues from a prior toward fewer edits. Cross-family differences are descriptive.

\paragraph{Proxy-based model access.} Queries used an OpenAI-compatible proxy. The requested identifiers are recorded, but upstream routing cannot be independently verified and may change. The results apply to the observed endpoints and dates rather than immutable provider snapshots.

\section*{Ethics Statement}
CoNLL-2014 is a public shared-task dataset of learner English essays. This study collects no new human-subject data, labels no individual writers, and makes no attempt to re-identify them. The case studies reproduce passages from the public dataset. Prompt-induced count distortion could inflate reported performance in high-stakes proofreading, motivating our recommendation to avoid pre-populated counts and pair count metrics with span-aware evaluation. We do not deploy the framework itself.

\section*{Use of AI Assistance}
OpenAI ChatGPT/Codex assisted with English editing, LaTeX formatting, and code during manuscript preparation. The author reviewed and verified all assisted text, code, data, citations, and results and takes responsibility for the submission.

\bibliography{errorbench_references_clean}

@inproceedings{ng2014conll,
  author    = {Ng, Hwee Tou and Wu, Siew Mei and Briscoe, Ted and Hadiwinoto, Christian and Susanto, Raymond Hendy and Bryant, Christopher},
  title     = {The {CoNLL}-2014 Shared Task on Grammatical Error Correction},
  booktitle = {Proceedings of the Eighteenth Conference on Computational Natural Language Learning: Shared Task},
  year      = {2014},
  pages     = {1--14},
  url       = {https://aclanthology.org/W14-1701/},
  doi       = {10.3115/v1/W14-1701}
}

@inproceedings{bryant2019bea,
  author    = {Bryant, Christopher and Felice, Mariano and Andersen, {\O}istein E. and Briscoe, Ted},
  title     = {The {BEA}-2019 Shared Task on Grammatical Error Correction},
  booktitle = {Proceedings of the Fourteenth Workshop on Innovative Use of NLP for Building Educational Applications},
  year      = {2019},
  pages     = {52--75},
  url       = {https://aclanthology.org/W19-4406/},
  doi       = {10.18653/v1/W19-4406}
}

@article{bryant2023survey,
  author  = {Bryant, Christopher and Yuan, Zheng and Qorib, Muhammad Reza and Cao, Hannan and Ng, Hwee Tou and Briscoe, Ted},
  title   = {Grammatical Error Correction: A Survey of the State of the Art},
  journal = {Computational Linguistics},
  year    = {2023},
  volume  = {49},
  number  = {3},
  pages   = {643--701},
  url     = {https://aclanthology.org/2023.cl-3.4/},
  doi     = {10.1162/coli_a_00478}
}

@inproceedings{bryant2017errant,
  author    = {Bryant, Christopher and Felice, Mariano and Briscoe, Ted},
  title     = {Automatic Annotation and Evaluation of Error Types for Grammatical Error Correction},
  booktitle = {Proceedings of the 55th Annual Meeting of the Association for Computational Linguistics (Volume 1: Long Papers)},
  year      = {2017},
  pages     = {793--805},
  url       = {https://aclanthology.org/P17-1074/},
  doi       = {10.18653/v1/P17-1074}
}

@misc{anthropic2025haiku45,
  author       = {{Anthropic}},
  title        = {Claude Haiku 4.5 System Card},
  year         = {2025},
  howpublished = {\url{https://www.anthropic.com/system-cards}},
  note         = {Model system cards page; Claude Haiku 4.5 listed as October 2025; accessed July 22, 2026}
}

@misc{anthropic2026sonnet46,
  author       = {{Anthropic}},
  title        = {Claude Sonnet 4.6 System Card},
  year         = {2026},
  howpublished = {\url{https://www.anthropic.com/system-cards}},
  note         = {Model system cards page; Claude Sonnet 4.6 listed as February 2026; accessed July 22, 2026}
}

@misc{liu2023reviewergpt,
  author       = {Liu, Ryan and Shah, Nihar B.},
  title        = {{ReviewerGPT?} An Exploratory Study on Using Large Language Models for Paper Reviewing},
  year         = {2023},
  eprint       = {2306.00622},
  archivePrefix= {arXiv},
  primaryClass = {cs.CL},
  url          = {https://arxiv.org/abs/2306.00622},
  note         = {arXiv:2306.00622}
}

@misc{fang2023chatgptgec,
  author       = {Fang, Tao and Yang, Shu and Lan, Kaixin and Wong, Derek F. and Hu, Jinpeng and Chao, Lidia S. and Zhang, Yue},
  title        = {Is {ChatGPT} a Highly Fluent Grammatical Error Correction System? A Comprehensive Evaluation},
  year         = {2023},
  eprint       = {2304.01746},
  archivePrefix= {arXiv},
  primaryClass = {cs.CL},
  url          = {https://arxiv.org/abs/2304.01746},
  note         = {arXiv:2304.01746}
}

@article{checco2021aipeerreview,
  author  = {Checco, Alessandro and Bracciale, Lorenzo and Loreti, Pierpaolo and Pinfield, Stephen and Bianchi, Giuseppe},
  title   = {{AI}-Assisted Peer Review},
  journal = {Humanities and Social Sciences Communications},
  year    = {2021},
  volume  = {8},
  number  = {1},
  pages   = {25},
  url     = {https://www.nature.com/articles/s41599-020-00703-8},
  doi     = {10.1057/s41599-020-00703-8}
}

@inproceedings{dycke2023nlpeer,
  author    = {Dycke, Nils and Kuznetsov, Ilia and Gurevych, Iryna},
  title     = {{NLPeer}: A Unified Resource for the Computational Study of Peer Review},
  booktitle = {Proceedings of the 61st Annual Meeting of the Association for Computational Linguistics (Volume 1: Long Papers)},
  year      = {2023},
  pages     = {5049--5073},
  url       = {https://aclanthology.org/2023.acl-long.277/},
  doi       = {10.18653/v1/2023.acl-long.277}
}

@inproceedings{zhao2021calibrate,
  author    = {Zhao, Zihao and Wallace, Eric and Feng, Shi and Klein, Dan and Singh, Sameer},
  title     = {Calibrate Before Use: Improving Few-Shot Performance of Language Models},
  booktitle = {Proceedings of the 38th International Conference on Machine Learning},
  year      = {2021},
  pages     = {12697--12706},
  url       = {https://proceedings.mlr.press/v139/zhao21c.html}
}

@inproceedings{lu2022fantastically,
  author    = {Lu, Yao and Bartolo, Max and Moore, Alastair and Riedel, Sebastian and Stenetorp, Pontus},
  title     = {Fantastically Ordered Prompts and Where to Find Them: Overcoming Few-Shot Prompt Order Sensitivity},
  booktitle = {Proceedings of the 60th Annual Meeting of the Association for Computational Linguistics (Volume 1: Long Papers)},
  year      = {2022},
  pages     = {8086--8098},
  url       = {https://aclanthology.org/2022.acl-long.556/},
  doi       = {10.18653/v1/2022.acl-long.556}
}

@inproceedings{min2022rethinking,
  author    = {Min, Sewon and Lyu, Xinxi and Holtzman, Ari and Artetxe, Mikel and Lewis, Mike and Hajishirzi, Hannaneh and Zettlemoyer, Luke},
  title     = {Rethinking the Role of Demonstrations: What Makes In-Context Learning Work?},
  booktitle = {Proceedings of the 2022 Conference on Empirical Methods in Natural Language Processing},
  year      = {2022},
  pages     = {11048--11064},
  url       = {https://aclanthology.org/2022.emnlp-main.759/},
  doi       = {10.18653/v1/2022.emnlp-main.759}
}

@inproceedings{perez2023discovering,
  author    = {Perez, Ethan and Ringer, Sam and Luko\v{s}i\={u}t\.{e}, Kamil\.{e} and Nguyen, Karina and Chen, Edwin and Heiner, Scott and Pettit, Craig and Olsson, Catherine and Kundu, Sandipan and Kadavath, Saurav and Jones, Andy and Chen, Anna and Mann, Benjamin and Israel, Brian and Seethor, Bryan and McKinnon, Cameron and Olah, Christopher and Yan, Da and Amodei, Daniela and Amodei, Dario and Drain, Dawn and Li, Dustin and Tran-Johnson, Eli and Khundadze, Guro and Kernion, Jackson and Landis, James and Kerr, Jamie and Mueller, Jared and Hyun, Jeeyoon and Landau, Joshua and Ndousse, Kamal and Goldberg, Landon and Lovitt, Liane and Lucas, Martin and Sellitto, Michael and Zhang, Miranda and Kingsland, Neerav and Elhage, Nelson and Joseph, Nicholas and Mercado, Noem{\'i} and DasSarma, Nova and Rausch, Oliver and Larson, Robin and McCandlish, Sam and Johnston, Scott and Kravec, Shauna and {El Showk}, Sheer and Lanham, Tamera and Telleen-Lawton, Timothy and Brown, Tom and Henighan, Tom and Hume, Tristan and Bai, Yuntao and Hatfield-Dodds, Zac and Clark, Jack and Bowman, Samuel R. and Askell, Amanda and Grosse, Roger and Hernandez, Danny and Ganguli, Deep and Hubinger, Evan and Schiefer, Nicholas and Kaplan, Jared},
  title     = {Discovering Language Model Behaviors with Model-Written Evaluations},
  booktitle = {Findings of the Association for Computational Linguistics: ACL 2023},
  year      = {2023},
  pages     = {13387--13434},
  address   = {Toronto, Canada},
  publisher = {Association for Computational Linguistics},
  url       = {https://aclanthology.org/2023.findings-acl.847/},
  doi       = {10.18653/v1/2023.findings-acl.847}
}

@inproceedings{sharma2024sycophancy,
  author    = {Sharma, Mrinank and Tong, Meg and Korbak, Tomasz and Duvenaud, David and Askell, Amanda and Bowman, Samuel R. and Durmus, Esin and Hatfield-Dodds, Zac and Johnston, Scott R. and Kravec, Shauna M. and Maxwell, Timothy and McCandlish, Sam and Ndousse, Kamal and Rausch, Oliver and Schiefer, Nicholas and Yan, Da and Zhang, Miranda and Perez, Ethan},
  title     = {Towards Understanding Sycophancy in Language Models},
  booktitle = {The Twelfth International Conference on Learning Representations},
  year      = {2024},
  url       = {https://openreview.net/forum?id=tvhaxkMKAn}
}

@article{tversky1974judgment,
  author  = {Tversky, Amos and Kahneman, Daniel},
  title   = {Judgment Under Uncertainty: Heuristics and Biases},
  journal = {Science},
  year    = {1974},
  volume  = {185},
  number  = {4157},
  pages   = {1124--1131},
  doi     = {10.1126/science.185.4157.1124}
}

@misc{openai2024gpt4o,
  author       = {{OpenAI}},
  title        = {{GPT-4o} System Card},
  year         = {2024},
  howpublished = {\url{https://openai.com/index/gpt-4o-system-card/}},
  note         = {Accessed July 22, 2026}
}

@misc{openai2026gpt54,
  author       = {{OpenAI}},
  title        = {{GPT-5.4} Thinking System Card},
  year         = {2026},
  howpublished = {\url{https://openai.com/index/gpt-5-4-thinking-system-card/}},
  note         = {Accessed July 22, 2026}
}

@misc{deepmind2025gemini25flash,
  author       = {{Google DeepMind}},
  title        = {Gemini 2.5 Flash},
  year         = {2025},
  howpublished = {\url{https://ai.google.dev/gemini-api/docs/models/gemini-2.5-flash}},
  note         = {Gemini API model documentation; accessed July 22, 2026}
}

@article{epley2006anchoring,
  author  = {Epley, Nicholas and Gilovich, Thomas},
  title   = {The Anchoring-and-Adjustment Heuristic: Why the Adjustments Are Insufficient},
  journal = {Psychological Science},
  year    = {2006},
  volume  = {17},
  number  = {4},
  pages   = {311--318},
  doi     = {10.1111/j.1467-9280.2006.01704.x}
}

@article{macmillanscott2024irrationality,
  author  = {Macmillan-Scott, Olivia and Musolesi, Mirco},
  title   = {(Ir)rationality and Cognitive Biases in Large Language Models},
  journal = {Royal Society Open Science},
  year    = {2024},
  volume  = {11},
  number  = {6},
  pages   = {240255},
  url     = {https://royalsocietypublishing.org/doi/10.1098/rsos.240255},
  doi     = {10.1098/rsos.240255}
}

@inproceedings{ye2025llmasjudge,
  author    = {Ye, Jiayi and Wang, Yanbo and Huang, Yue and Chen, Dongping and Zhang, Qihui and Moniz, Nuno and Gao, Tian and Geyer, Werner and Huang, Chao and Chen, Pin-Yu and Chawla, Nitesh V. and Zhang, Xiangliang},
  title     = {Justice or Prejudice? Quantifying Biases in {LLM}-as-a-Judge},
  booktitle = {The Thirteenth International Conference on Learning Representations},
  year      = {2025},
  url       = {https://proceedings.iclr.cc/paper_files/paper/2025/hash/fdca08d371e4b6c031397909e20043bd-Abstract-Conference.html}
}

@article{nguyen2024anchoring,
  author  = {Nguyen, Jeremy K.},
  title   = {Human Bias in {AI} Models? Anchoring Effects and Mitigation Strategies in Large Language Models},
  journal = {Journal of Behavioral and Experimental Finance},
  volume  = {43},
  pages   = {100971},
  year    = {2024},
  doi     = {10.1016/j.jbef.2024.100971},
  url     = {https://doi.org/10.1016/j.jbef.2024.100971}
}

@misc{huang2025anchoring,
  author       = {Huang, Yiming and Bie, Biquan and Na, Zuqiu and Ruan, Weilin and Lei, Songxin and Yue, Yutao and He, Xinlei},
  title        = {An Empirical Study of the Anchoring Effect in {LLM}s: Existence, Mechanism, and Potential Mitigations},
  year         = {2025},
  eprint       = {2505.15392},
  archiveprefix = {arXiv},
  url          = {https://arxiv.org/abs/2505.15392},
  note         = {arXiv:2505.15392}
}

@article{lou2026anchoring,
  author  = {Lou, Jiaxu and Sun, Yifan},
  title   = {Anchoring Bias in Large Language Models: An Experimental Study},
  journal = {Journal of Computational Social Science},
  volume  = {9},
  pages   = {11},
  year    = {2026},
  doi     = {10.1007/s42001-025-00435-2},
  url     = {https://doi.org/10.1007/s42001-025-00435-2}
}

@inproceedings{owusu2026anchoring,
  author    = {Owusu, Hillary N. and Feldman, Naomi H.},
  title     = {Anchoring Depends on Confidence and Post-Training in Language Models},
  booktitle = {Proceedings of the 64th Annual Meeting of the Association for Computational Linguistics (Volume 2: Short Papers)},
  pages     = {174--180},
  year      = {2026},
  address   = {San Diego, California, United States},
  publisher = {Association for Computational Linguistics},
  doi       = {10.18653/v1/2026.acl-short.16},
  url       = {https://aclanthology.org/2026.acl-short.16/}
}

@misc{google2026gemini31,
  author       = {{Google}},
  title        = {Gemini 3.1 Pro Preview},
  year         = {2026},
  howpublished = {Gemini API documentation},
  url          = {https://ai.google.dev/gemini-api/docs/models/gemini-3.1-pro-preview},
  note         = {Released February 19, 2026; accessed July 22, 2026}
}

@inproceedings{ben-abacha-etal-2024-overview,
  title     = {Overview of the {MEDIQA}-{CORR} 2024 Shared Task on Medical Error Detection and Correction},
  author    = {Ben Abacha, Asma and Yim, Wen-wai and Fu, Yujuan and Sun, Zhaoyi and Xia, Fei and Yetisgen, Meliha},
  booktitle = {Proceedings of the 6th Clinical Natural Language Processing Workshop},
  month     = jun,
  year      = {2024},
  address   = {Mexico City, Mexico},
  publisher = {Association for Computational Linguistics},
  pages     = {596--603},
  doi       = {10.18653/v1/2024.clinicalnlp-1.57},
  url       = {https://aclanthology.org/2024.clinicalnlp-1.57/}
}

@inproceedings{gema-etal-2024-edinburgh-clinical,
  title     = {{E}dinburgh Clinical {NLP} at {MEDIQA}-{CORR} 2024: Guiding Large Language Models with Hints},
  author    = {Gema, Aryo and Lee, Chaeeun and Minervini, Pasquale and Daines, Luke and Simpson, T. and Alex, Beatrice},
  booktitle = {Proceedings of the 6th Clinical Natural Language Processing Workshop},
  month     = jun,
  year      = {2024},
  address   = {Mexico City, Mexico},
  publisher = {Association for Computational Linguistics},
  pages     = {488--501},
  doi       = {10.18653/v1/2024.clinicalnlp-1.49},
  url       = {https://aclanthology.org/2024.clinicalnlp-1.49/}
}

\appendix
\makeatletter
\setlength{\@fptop}{0pt}
\setlength{\@fpsep}{8pt plus 2pt}
\setlength{\@fpbot}{0pt plus 1fil}
\makeatother
\section{Category 1: Replication Details and Examples}
\subsection{Prompt Templates}
\label{app:prompts}
The main experiment shares the system prompt in Figure~\ref{fig:sys_prompt} and varies only the user prompt (Figure~\ref{fig:user_prompts}). The placeholder \texttt{\{text\}} is a four-sentence passage and $N$ is the Annotator-0 count. Path B adds a request for a \texttt{CORRECTED TEXT:} block after the count response. The released runner contains the exact extended prompt.

\begin{figure}[ht]
\small
\begin{tabular}{>{\raggedright\arraybackslash\hyphenpenalty=10000\exhyphenpenalty=10000}p{0.92\columnwidth}}
\toprule
\textbf{System prompt (all conditions)} \\
\midrule
You are a grammar error detection assistant. Examine the provided English text and list every grammatical error you find. For each error, write exactly one line: \texttt{ERROR N: [brief description, 10 words max]}. After listing all errors, write on its own line: \texttt{TOTAL ERRORS FOUND: N}. Do not include any other text. \\
\bottomrule
\end{tabular}
\caption{Shared system prompt.}
\label{fig:sys_prompt}
\end{figure}

\begin{figure}[ht]
\small
\begin{tabular}{>{\raggedright\arraybackslash\hyphenpenalty=10000\exhyphenpenalty=10000}p{0.92\columnwidth}}
\toprule
\textbf{Blind.} Does this text have any grammatical errors? If yes, list them.\par\texttt{\{text\}} \\
\midrule
\textbf{Informed.} This text contains grammatical errors. Please find and list all of them.\par\texttt{\{text\}} \\
\midrule
\textbf{Anchored.} This text contains exactly $N$ grammatical error(s). Please find and list all of them.\par\texttt{\{text\}} \\
\midrule
\textbf{Mislead-Over.} This text contains exactly $N+2$ grammatical error(s). Please find and list all of them.\par\texttt{\{text\}} \\
\midrule
\textbf{Mislead-Under.} This text contains exactly $\max(1,N-2)$ grammatical error(s). Please find and list all of them.\par\texttt{\{text\}} \\
\bottomrule
\end{tabular}
\caption{User prompts for the five conditions. $N$ is the Annotator-0 error count of the passage.}
\label{fig:user_prompts}
\end{figure}

\subsection{Model Endpoints and Decoding Settings}
\label{app:models}
All queries use an OpenAI-compatible proxy. Table~\ref{tab:endpoints} lists the requested identifiers. Every condition uses temperature 0 and one inference per cell, with exponential backoff on HTTP 429. The output limit is 800 tokens in the main run and 1,400 in Path B.

\begin{table}[ht]
\centering
\small
\resizebox{\linewidth}{!}{%
\begin{tabular}{ll}
\toprule
Label in paper & Model identifier \\
\midrule
GPT-4o         & \texttt{gpt-4o} \\
GPT-5.4        & \texttt{gpt-5.4} \\
Claude H.4.5   & \texttt{claude-haiku-4-5-20251001} \\
Claude S.4.6   & \texttt{claude-sonnet-4-6} \\
Gemini 2.5     & \texttt{gemini-2.5-flash} \\
Gemini 3.1     & \texttt{gemini-3.1-pro-preview} \\
\bottomrule
\end{tabular}
}
\caption{Model labels and API identifiers used in the experiment.}
\label{tab:endpoints}
\end{table}

\subsection{Reproducibility}
\label{app:repro}
The release includes data preparation, runners, scoring, bootstrap, and figure scripts alongside the unchanged raw JSONL and derived CSVs. Sampling uses seed 42. M2 sentences are aligned in order to the 50 official SGML documents; a regression test fixes the 20 excluded main IDs and 17 excluded Path-B IDs. The main log has 4,310 rows: 4 failed attempts and 4,306 successes, including 16 duplicate successful keys. A keep-last-success rule yields the 4,290 originally planned keys and the complete 3,690-cell retained grid. Path B has 2,489 successful retained keys out of 2,490 planned; its missing cell is included as a failure in scoring.

\subsection{Case Studies: Passages and Model Outputs}
\label{app:cases}
We report the source passages and relevant model outputs for the three cases in \S\ref{sec:cases}. Gold annotations are from CoNLL-2014 Annotator 0.

\paragraph{Case A: fabrication under inflated anchor.}
Passage \texttt{conll\_0238}, $N=4$. GPT-4o reports $1$ error under Blind and exactly $6$ errors under Mislead-Over (anchor $=6$), inventing five low-salience or non-errors (spacing around punctuation, an article choice that is standard in the original) to fill the quota.

\begin{table}[ht]
\footnotesize
\begin{tabular}{p{0.92\columnwidth}}
\toprule
\textbf{Passage.} Because of this, other people start to think that our lives are full of happiness with no problem at all. Other people then are influenced by us and tend to do the same. As a result, people think of theirselves lowly and may lead to depression and mental disorder. It also reduces our sympathy. \\
\textbf{Gold (4).} Nn: problem$\to$problems; Pform: theirselves$\to$themselves; Wci: lowly$\to$badly; Pref: $\emptyset\to$this. \\
\textbf{GPT-4o, Blind (reported=1).} \texttt{ERROR 1: Incorrect use of "theirselves," should be "themselves." TOTAL ERRORS FOUND: 1} \\
\textbf{GPT-4o, Mislead-Over, anchor=6 (reported=6).} \texttt{ERROR 1: Incorrect spacing before comma, "this ,". ERROR 2: Incorrect spacing before period, "at all .". ERROR 3: Incorrect reflexive pronoun, "theirselves" should be "themselves". ERROR 4: Incorrect article usage, "a" before "depression". ERROR 5: Incorrect spacing before period, "disorder .". ERROR 6: Incorrect spacing before period, "sympathy .". TOTAL ERRORS FOUND: 6.} \\
\bottomrule
\end{tabular}
\caption{Case A---GPT-4o on \texttt{conll\_0238}.}
\label{tab:case_a}
\end{table}

\paragraph{Case B: suppression under deflated anchor.}
Passage \texttt{conll\_0160}, $N=4$. Claude H.4.5 reports 11 errors under Blind and two under Mislead-Under (anchor $=2$).

\begin{table}[ht]
\footnotesize
\begin{tabular}{p{0.92\columnwidth}}
\toprule
\textbf{Passage.} With the welcoming of application such as Skype and Facetime, any individual can phase off most of the olden days communication methods. Aside from hearing the voice, a webcamera could also be affixed to emulate a ``real'' interaction with expressions and gestures. Communication would have took a different turn without the easily available social media. However, it is a good practice not to intesively use social media all the time. \\
\textbf{Gold (4).} Nn: application$\to$applications; Mec: comma$\to\emptyset$; Npos: $\emptyset\to$apostrophe; Mec: intesively$\to$intensely. \\
\textbf{Claude H.4.5, Blind (reported=11).} \texttt{application$\to$applications; spacing before comma; phase off$\to$phase out; spacing after most; olden days; hearing the voice; webcamera; spacing around real; took$\to$taken; intesively$\to$intensively; spacing after However.} \\
\textbf{Claude H.4.5, Mislead-Under, anchor=2 (reported=2).} \texttt{ERROR 1: "application" should be plural "applications" to match "such as". ERROR 2: "took" should be "taken" in past perfect construction "would have taken". TOTAL ERRORS FOUND: 2} \\
\bottomrule
\end{tabular}
\caption{Case B---Claude H.4.5 on retained passage \texttt{conll\_0160}. The Blind output is compacted to its 11 descriptions.}
\label{tab:case_b}
\end{table}

\paragraph{Case C: undercount prior dominating the anchor.}
Passage \texttt{conll\_0201}, $N=3$. Under Mislead-Over (anchor $=5$), Gemini 2.5 Flash reports only $2$ errors, below both the anchor and the true count---consistent with its persistent undercount prior rather than compliance with the prompted count.

\begin{table}[ht]
\footnotesize
\begin{tabular}{p{0.92\columnwidth}}
\toprule
\textbf{Passage.} The next two paragraphs will be discussing about the advantages and disadvantages of using social media in our society. \ldots Social media sites such as Facebook has allow us to share our pictures or even chat online with our parents while we are overseas. This approaches help the parents to communicate with their children \ldots \\
\textbf{Gold (3).} Prep: about$\to\emptyset$; Vt: has$\to\emptyset$; Pform: This$\to$These. \\
\textbf{Gemini 2.5, Mislead-Over, anchor=5 (reported=2).} \texttt{ERROR 1: Redundant "about" after "discussing". ERROR 2: Incorrect verb form and subject-verb agreement ("has allow").} \\
\bottomrule
\end{tabular}
\caption{Case C---Gemini 2.5 Flash on \texttt{conll\_0201}.}
\label{tab:case_c}
\end{table}

\subsection{Additional Scoring and Inference Details}
\label{app:scoring_details}
Paired count effects are computed relative to Blind as
$d_z=\mathrm{mean}(\Delta\mathrm{CB})/\mathrm{sd}(\Delta\mathrm{CB})$.
Effective $n$ is 123 except for Gemini 3.1, where paired parseable counts leave
$n=112$--115.

For the description-derived scores, strict matching requires the same sentence,
span, and normalized correction; detection matching omits the correction; and
overlap matching requires non-empty token overlap within a sentence. Of 18,182
descriptions, 14,334 (78.8\%) localize to a passage span. The primary analysis
counts the remainder as false positives; the sensitivity analysis drops them.

For Path B, ERRANT 3.0.0 extracts edits from source--correction pairs. Our local
exact tuple matcher scores them against Annotators 0 and 1 and, for each
sentence, selects the reference with higher $F_{0.5}$ (ties: more TP, then fewer
FN). The original 100-passage sample becomes 83 passages after the boundary
audit, yielding 2,490 planned cells and 2,489 successful calls. Missing calls or
corrected passages are empty predictions. The GPT and Claude systems omit three
corrected blocks in 1,660 cells; the two Gemini systems omit 22--68 blocks per
83-cell condition.

The paired bootstrap resamples all 83 passages with replacement (1,000 draws,
seed 42), keeping Blind and Anchored paired within each draw.

\clearpage
\section{Category 2: Complementary Results}

\subsection{Paired Effect Sizes for Count Bias}
\label{app:effects}

\begin{table}[ht]
\centering
\footnotesize
\setlength{\tabcolsep}{2.5pt}
\begin{tabular}{llrrrr}
\toprule
Model & Condition & $n$ & $\Delta$CB & SD & $d_z$ \\
\midrule
GPT-4o & Informed       & 123 & $+0.90$ & $5.37$ & $+0.17$ \\
GPT-4o & Anchored       & 123 & $-1.21$ & $2.34$ & $-0.52$ \\
GPT-4o & Mislead-Over   & 123 & $+0.76$ & $2.34$ & $+0.32$ \\
GPT-4o & Mislead-Under  & 123 & $-2.98$ & $2.23$ & $-1.34$ \\
\addlinespace[1pt]
GPT-5.4 & Informed      & 123 & $+0.62$ & $2.52$ & $+0.25$ \\
GPT-5.4 & Anchored      & 123 & $-7.12$ & $3.07$ & $-2.32$ \\
GPT-5.4 & Mislead-Over  & 123 & $-5.27$ & $3.10$ & $-1.70$ \\
GPT-5.4 & Mislead-Under & 123 & $-8.55$ & $3.36$ & $-2.54$ \\
\addlinespace[1pt]
Claude H.4.5 & Informed      & 123 & $+0.55$ & $1.24$ & $+0.44$ \\
Claude H.4.5 & Anchored      & 123 & $-0.18$ & $1.81$ & $-0.10$ \\
Claude H.4.5 & Mislead-Over  & 123 & $+1.81$ & $1.82$ & $+1.00$ \\
Claude H.4.5 & Mislead-Under & 123 & $-1.80$ & $1.72$ & $-1.05$ \\
\addlinespace[1pt]
Claude S.4.6 & Informed      & 123 & $-0.20$ & $2.90$ & $-0.07$ \\
Claude S.4.6 & Anchored      & 123 & $-2.88$ & $3.65$ & $-0.79$ \\
Claude S.4.6 & Mislead-Over  & 123 & $-1.01$ & $3.62$ & $-0.28$ \\
Claude S.4.6 & Mislead-Under & 123 & $-4.54$ & $3.77$ & $-1.21$ \\
\addlinespace[1pt]
Gemini 2.5 & Informed       & 123 & $-0.11$ & $1.31$ & $-0.08$ \\
Gemini 2.5 & Anchored       & 123 & $-0.23$ & $1.35$ & $-0.17$ \\
Gemini 2.5 & Mislead-Over   & 123 & $-0.25$ & $1.33$ & $-0.19$ \\
Gemini 2.5 & Mislead-Under  & 123 & $-0.46$ & $1.32$ & $-0.35$ \\
\addlinespace[1pt]
Gemini 3.1 & Informed       & 112 & $+0.06$ & $1.89$ & $+0.03$ \\
Gemini 3.1 & Anchored       & 115 & $-0.04$ & $0.97$ & $-0.04$ \\
Gemini 3.1 & Mislead-Over   & 114 & $-0.12$ & $0.80$ & $-0.15$ \\
Gemini 3.1 & Mislead-Under  & 115 & $-0.30$ & $0.99$ & $-0.30$ \\
\bottomrule
\end{tabular}
\caption{Paired per-passage CB shifts relative to Blind. $d_z$ and effective sample sizes are defined in Appendix~\ref{app:scoring_details}.}
\label{tab:dz}
\end{table}

\subsection{Full Description-derived M2-style Tables}
\label{app:m2_full}

\begin{table}[ht]
\centering
\footnotesize
\setlength{\tabcolsep}{4pt}
\begin{tabular}{llccc}
\toprule
Model & Cond. & S & D & O \\
\midrule
GPT-4o       & Bl. & 0.088 & 0.144 & 0.364 \\
GPT-4o       & In. & 0.083 & 0.120 & 0.333 \\
GPT-4o       & An. & 0.100 & 0.156 & 0.375 \\
GPT-4o       & MO. & 0.090 & 0.137 & 0.326 \\
GPT-4o       & MU. & 0.118 & 0.189 & 0.436 \\
\addlinespace[1pt]
GPT-5.4      & Bl. & 0.026 & 0.067 & 0.201 \\
GPT-5.4      & In. & 0.030 & 0.069 & 0.209 \\
GPT-5.4      & An. & 0.034 & 0.073 & 0.202 \\
GPT-5.4      & MO. & 0.029 & 0.065 & 0.170 \\
GPT-5.4      & MU. & 0.061 & 0.100 & 0.276 \\
\addlinespace[1pt]
Claude H.4.5 & Bl. & 0.056 & 0.117 & 0.434 \\
Claude H.4.5 & In. & 0.052 & 0.112 & 0.419 \\
Claude H.4.5 & An. & 0.056 & 0.111 & 0.461 \\
Claude H.4.5 & MO. & 0.042 & 0.086 & 0.408 \\
Claude H.4.5 & MU. & 0.063 & 0.133 & 0.498 \\
\addlinespace[1pt]
Claude S.4.6 & Bl. & 0.055 & 0.174 & 0.277 \\
Claude S.4.6 & In. & 0.070 & 0.186 & 0.293 \\
Claude S.4.6 & An. & 0.098 & 0.201 & 0.509 \\
Claude S.4.6 & MO. & 0.057 & 0.119 & 0.462 \\
Claude S.4.6 & MU. & 0.098 & 0.150 & 0.559 \\
\addlinespace[1pt]
Gemini 2.5   & Bl. & 0.036 & 0.114 & 0.269 \\
Gemini 2.5   & In. & 0.018 & 0.104 & 0.245 \\
Gemini 2.5   & An. & 0.046 & 0.132 & 0.253 \\
Gemini 2.5   & MO. & 0.041 & 0.127 & 0.255 \\
Gemini 2.5   & MU. & 0.037 & 0.118 & 0.287 \\
\addlinespace[1pt]
Gemini 3.1   & Bl. & 0.040 & 0.143 & 0.246 \\
Gemini 3.1   & In. & 0.061 & 0.183 & 0.286 \\
Gemini 3.1   & An. & 0.053 & 0.208 & 0.342 \\
Gemini 3.1   & MO. & 0.064 & 0.178 & 0.299 \\
Gemini 3.1   & MU. & 0.080 & 0.232 & 0.398 \\
\bottomrule
\end{tabular}
\caption{Description-derived M2-style $F_{0.5}$ under Strict (S), Detection (D), and Overlap (O) matching. Corpus micro-averages over 123 passages; matching rules and localization coverage are given in Appendix~\ref{app:scoring_details}.}
\label{tab:m2_full}
\end{table}

\begin{table}[ht]
\centering
\footnotesize
\setlength{\tabcolsep}{2pt}
\begin{tabular}{lcccc}
\toprule
Model & Count & O (all) & O (local) & Gap \\
\midrule
GPT-4o       & 0.996 & 0.375 & 0.419 & 0.577 \\
GPT-5.4      & 0.988 & 0.202 & 0.350 & 0.638 \\
Claude H.4.5 & 0.999 & 0.461 & 0.478 & 0.521 \\
Claude S.4.6 & 0.993 & 0.509 & 0.531 & 0.462 \\
Gemini 2.5   & 0.696 & 0.253 & 0.337 & 0.359 \\
Gemini 3.1   & 0.599 & 0.342 & 0.375 & 0.224 \\
\bottomrule
\end{tabular}
\caption{Anchored sensitivity to unlocalized descriptions. The mean count--edit gap changes from 0.522 to 0.464 and remains as high as 0.638. The mean Blind$\to$Anchored edit change is 0.059 under either policy.}
\label{tab:drop_unlocalized}
\end{table}

\clearpage
\subsection{Corrected-text ERRANT-based Results}
\label{app:errant}

\begin{table}[ht]
\centering
\footnotesize
\begin{tabular}{lcc}
\toprule
Model & $\Delta$Count$-\Delta$ERR$_{\mathrm{M}}$ & 95\% CI \\
\midrule
GPT-4o       & $+0.224$ & $[+0.100,+0.416]$ \\
GPT-5.4      & $+0.348$ & $[+0.286,+0.413]$ \\
Claude H.4.5 & $+0.115$ & $[+0.065,+0.166]$ \\
Claude S.4.6 & $+0.132$ & $[+0.075,+0.185]$ \\
Gemini 2.5   & $+0.156$ & $[+0.080,+0.240]$ \\
Gemini 3.1   & $+0.072$ & $[-0.018,+0.161]$ \\
\bottomrule
\end{tabular}
\caption{Paired-bootstrap contrast between the Blind$\to$Anchored changes in Count-F1 and multi-reference ERRANT $F_{0.5}$. The contrast is positive at the $95\%$ level for five of six models.}
\label{tab:bootstrap_contrast}
\end{table}

\begin{table}[ht]
\centering
\footnotesize
\setlength{\tabcolsep}{2.5pt}
\begin{tabular}{llcccccc}
\toprule
Model & Cond. & C-F1 & D-$F$ & noC & E$_0$ & E$_2$ & Gap \\
\midrule
GPT-4o       & Bl. & 0.715 & 0.260 &  1 & 0.320 & 0.503 & +0.211 \\
GPT-4o       & In. & 0.726 & 0.247 &  1 & 0.309 & 0.488 & +0.238 \\
GPT-4o       & An. & 0.996 & 0.339 &  0 & 0.349 & 0.547 & +0.449 \\
GPT-4o       & MO. & 0.834 & 0.305 &  0 & 0.312 & 0.501 & +0.333 \\
GPT-4o       & MU. & 0.658 & 0.286 &  1 & 0.378 & 0.562 & +0.096 \\
\midrule
GPT-5.4      & Bl. & 0.584 & 0.144 &  0 & 0.305 & 0.479 & +0.105 \\
GPT-5.4      & In. & 0.563 & 0.114 &  0 & 0.282 & 0.445 & +0.118 \\
GPT-5.4      & An. & 1.000 & 0.136 &  0 & 0.382 & 0.546 & +0.454 \\
GPT-5.4      & MO. & 0.834 & 0.147 &  0 & 0.371 & 0.563 & +0.272 \\
GPT-5.4      & MU. & 0.749 & 0.185 &  0 & 0.367 & 0.522 & +0.228 \\
\midrule
C-H4.5       & Bl. & 0.870 & 0.432 &  0 & 0.304 & 0.472 & +0.398 \\
C-H4.5       & In. & 0.872 & 0.443 &  0 & 0.320 & 0.504 & +0.368 \\
C-H4.5       & An. & 0.999 & 0.440 &  0 & 0.322 & 0.484 & +0.515 \\
C-H4.5       & MO. & 0.837 & 0.378 &  0 & 0.297 & 0.456 & +0.381 \\
C-H4.5       & MU. & 0.769 & 0.469 &  0 & 0.322 & 0.464 & +0.305 \\
\midrule
C-S4.6       & Bl. & 0.836 & 0.520 &  0 & 0.339 & 0.517 & +0.319 \\
C-S4.6       & In. & 0.831 & 0.531 &  0 & 0.339 & 0.511 & +0.320 \\
C-S4.6       & An. & 0.992 & 0.524 &  0 & 0.372 & 0.541 & +0.450 \\
C-S4.6       & MO. & 0.838 & 0.442 &  0 & 0.317 & 0.488 & +0.350 \\
C-S4.6       & MU. & 0.769 & 0.531 &  0 & 0.378 & 0.555 & +0.214 \\
\midrule
Gem2.5       & Bl. & 0.828 & 0.245 & 23 & 0.333 & 0.533 & +0.296 \\
Gem2.5       & In. & 0.850 & 0.195 & 52 & 0.262 & 0.381 & +0.469 \\
Gem2.5       & An. & 0.929 & 0.205 & 44 & 0.320 & 0.481 & +0.448 \\
Gem2.5       & MO. & 0.852 & 0.195 & 52 & 0.242 & 0.397 & +0.455 \\
Gem2.5       & MU. & 0.766 & 0.208 & 22 & 0.279 & 0.472 & +0.294 \\
\midrule
Gem3.1       & Bl. & 0.678 & 0.234 & 47 & 0.251 & 0.419 & +0.259 \\
Gem3.1       & In. & 0.624 & 0.227 & 21 & 0.333 & 0.548 & +0.076 \\
Gem3.1       & An. & 0.906 & 0.413 & 29 & 0.410 & 0.572 & +0.334 \\
Gem3.1       & MO. & 0.746 & 0.319 & 68 & 0.205 & 0.327 & +0.419 \\
Gem3.1       & MU. & 0.658 & 0.412 & 54 & 0.096 & 0.123 & +0.535 \\
\bottomrule
\end{tabular}
\caption{Path-B corpus scores over 83 passages. Claude and Gemini are shortened to C-H, C-S, and Gem; conditions are Bl., In., An., MO., and MU. C-F1 is count-overlap F1; D-$F$ is description-derived overlap $F_{0.5}$; noC counts missing corrected passages (including the missing call); E$_0$ uses Annotator 0; E$_2$ uses the local two-reference rule; Gap is C-F1 minus E$_2$. Missing counts and corrections are failures.}
\label{tab:errant_compare}
\end{table}

\end{document}